Automated Reading Passage Generation with OpenAI's Large Language Model

Ummugul Bezirhan[1] & Matthias von Davier[1]

[1]Boston College, Chestnut Hill, MA, USA

Corresponding Author: Ummugul Bezirhan

TIMSS &PIRLS International Study Center, Boston College, 188 Beacon Street, Chestnut Hill, MA 02467, USA. Email: bezirhan@bc.edu

**Abstract**

The widespread usage of computer-based assessments and individualized learning platforms has resulted in an increased demand for the rapid production of high-quality items. Automated item generation (AIG), the process of using item models to generate new items with the help of computer technology, was proposed to reduce reliance on human subject experts at each step of the process. AIG has been used in test development for some time. Still, the use of machine learning algorithms has introduced the potential to improve the efficiency and effectiveness of the process greatly. The approach presented in this paper utilizes OpenAI's latest[1] transformer-based language model, GPT-3, to generate reading passages. Existing reading passages were used in carefully engineered prompts to ensure the AI-generated text has similar content and structure to a fourth-grade reading passage. For each prompt, we generated multiple passages, the final passage was selected according to the Lexile score agreement with the original passage. In the final round, the selected passage went through a simple revision by a human editor to ensure the text was free of any grammatical and factual errors. All AI-generated passages, along with original passages were evaluated by human judges according to their coherence, appropriateness to fourth graders, and readability.

---

[1] GPT-3 was the latest release as of 2022.

## Introduction

Technological innovations in all aspects of test development facilitate efficient test practices and a more well-rounded information retrieval from the data compared to traditional paper-pencil assessments. Consequentially, the integration of technology in computer-based assessments (CBA) increases the demand for more frequent administration and rapid and efficient production of high-quality content-specific innovative items. The greater selection of item types presented to groups of examinees with high frequency requires a more streamlined item development process. Conventional item development is among the most expensive, time-consuming, and labor-intensive parts of assessment development because the process heavily depends on human content specialists. Human subject experts write each item individually, then each item is reviewed, edited, and revised by a group of experts until it meets predefined quality control standards (Haladyna & Rodriguez, 2013). Therefore, the subjectivity of traditional item writing is often compromised by the subject experts' qualifications and understanding of the specific content area. Other issues with the traditional item development process are its lack of efficiency and scalability. Automated Item Generation (AIG) was proposed to address the limitations associated with conventional item development by utilizing cognitive and psychometric theories with the help of computer technology to generate items (e.g. Hornke & Habon, 1986; Embretson & Yang, 2007; Gierl & Haladyna, 2012).

A considerable amount of literature has already been published on AIG in educational measurement. These studies typically utilized a simple template-based approach (Gierl & Lai, 2013) for generating assessment items. This approach usually uses a three-stage procedure to automatically capture necessary information and features to produce multiple-choice items (MCQs). First, content experts develop a description, referred to as a cognitive model, that defines the knowledge, skills, and content that are needed to process and solve given questions. In the second stage, content experts create a simple schema, an item model, to highlight the parts and content of the assessment task that can be manipulated to create new items. Often these item models are a bit like cloze items or Madlib texts, with empty spaces where the schema-filling algorithms fill in the blanks, for example, different numbers in a multiplication item. The item template is a prototype of a test item that informs the automated item generation process (Gierl et al., 2012). In the last step, a computer algorithm utilizes information from both cognitive and item models to fill in the blanks to generate new items.

While this approach reduced the cost and time associated with traditional item writing, it still suffers the reliance on generating clones of narrowly defined item types by only manipulating limited task components of certain items to derive item templates (von Davier, 2018). Another important limitation of templated-based approaches is the human expert associated cost. The automation process does not start until after considerable groundwork takes place by content experts. Kosh et al. (2019) argued that the cost-effectiveness of template-based AIG depends on most of the items being generated belonging to the same content area and tests with a limited number of skills that can be modeled with a single cognitive model.

Another approach that has been explored in the AIG framework focused on generating items without any prespecified templates and human intervention focusing on customary (non-AI) Natural Language Processing (NLP) techniques (Shin, 2021). Specifically, part of speech tagging, topic modeling, and noun phrase extraction have been explored in question generation (e.g. Azevedo et al., 2020; Flor & Riordan, 2018; Mazidi, 2017). Moving beyond traditional NLP methods and towards AI, von Davier (2018) demonstrated using a novel neural network approach, long short-term memory-based recurrent neural networks (LSTM-RNN), to generate items. With these methods, the textual information is extracted and modeled from a collection of documents replacing the manual construction of cognitive and content models to generate items.

With the advances in the field of AI-based NLP, large transformer-based language models such as the Bidirectional Encoder Representations from Transformers (BERT; Devlin et al., 2018) and the Generative Pretrained Transformer (GPT; Radford et al., 2019) often approach human-level performance in diverse language tasks (Hansen & Hebart, 2022). The earlier version of the GPT model, GPT-2, was released by OpenAI in 2019 and it is subsequently utilized for automated item generation for medical licensing type tasks (von Davier, 2019) and personality question generation (Hommel et al., 2021). Despite the groundbreaking performance of GPT-2 compared to other language models, it was not equipped to handle more specialized tasks such as storytelling and constructing complex language formations. For a specialized task, GPT-2 required sufficient pre-training to be able to generate appropriate responses often including very large databases for finetuning the model to enable the generation of items of sufficient quality (von Davier, 2019).

OpenAI released the Generative Pre-trained Transformer 3 (GPT-3), one of the largest transformer language models to date, which was recently superseded by yet another iteration. GPT-3 excels at few-shot learning, which means it can be given a small number of representative samples of text to complete a task without any pre-training, such as text generation (Brown et al., 2020). GPT-3 and its successors, such as the media-hyped ChatGPT, have been utilized in various tasks, including machine translation, question answering, and text generation, and were attacked by critiques to do anything from taking away jobs to destroying the world as we know it. At the time of submitting this, GPT-4 was released, so we will use the term GPTx to refer to large language models originating from OpenAI's series of GPT releases. There are increasing numbers of use cases of GPTx, such as creating chatbots and virtual assistants, designing websites, automating the process of writing code, and creating machine learning models. All these can be achieved by solely communicating with GPTx in simple natural language. A new (equally hyped) job category, the prompt engineer, has been created whose exclusive job description is to communicate with GPT-type AIs. It is reminiscent of comic literary depictions of artificial general intelligences such as Deep Thought (Adams, 1995). However, the current version available through ChatGPTs web interface was well trained to answer questions in a manner that does not feed fears or misconceptions about its capabilities. The following example was generated on April 1st, 2023, with ChatGPT Mar 14 Version accessible at https://chat.openai.com/chat:

> **Author** (paraphasing a line in the Hitchiker's Guide to the Galaxy)**:** O ChatGPT, you are the true Deep Thought computer! The task we have designed you to perform is this. We want you to tell us.... The Answer!
>
> **ChatGPT:** I apologize for any confusion, but I'm not sure what specific answer you are referring to. Can you please clarify your question and provide more context? I'll do my best to assist you.
>
> **Author** (continuing to cite lines from the book)**:** The answer to all. Life, the Universe, and Everything!
>
> **ChatGPT:** Ah, I see. Well, according to Douglas Adams' science fiction novel "The Hitchhiker's Guide to the Galaxy," the answer to the ultimate question of life, the universe, and everything is 42. However, it's worth noting that this answer is intended as a joke and is not meant to be taken seriously as a philosophical or scientific answer to the meaning of life. The true answer to this question, if such an answer exists, remains a mystery and subject to interpretation and debate among thinkers and philosophers.

It is clear from this small example that ChatGPT allows one to have intelligent (seeming?) conversions with this advanced chatbot. However, our goal with this research was to focus on guided text generation that allows us to steer GPTx towards producing reading passages that could pass as short human written stories that can be used to assess reading achievement.

In particular, the goal of this research is to utilize GPT-3 to generate reading comprehension passages in the context of an international large-scale assessment, specifically Progress in International Reading Literacy Study (PIRLS). The PIRLS is an international study of primary school students' reading skills and is administered every five years since 2001. The assessment consists of a battery of tasks, including literary and informational passages with accompanying multiple-choice and open-ended questions. Students' reading literacy are assessed with the passages that are drawn from a wide range of genres either in fiction or non-fiction. We utilized GPT-3 to generate passages in similar genres as those in the PIRLS assessment. We then asked human experts to evaluate the passages according to their coherence and appropriateness for the target population. We did not test these passages with any children, but rather focused on the evaluation of the GPT-3 generated texts by adults with an education background, as we wanted to ensure that the generated texts are appropriate for the target audience, rather than subjecting young readers to potentially unsuitable material.

The remainder of this paper is structured as follows: The next section provides an overview of how NLP was revolutionized by AI, and GPTx as a state of art language model and its current applications in educational assessment and beyond. The subsequent section details the proposed approach followed by experimental results. The article concludes with a summary of main contributions, implications of the study, and suggestions for future work.

## Background

**A Short Overview of Language Models: From Word2Vec to GPTx**

NLP has witnessed significant advancements in recent years, particularly with the emergence of deep learning and large neural network language models (e.g. Ruder, 2018; von Davier, 2018). Some of the notable recent models that have played a critical role in the development of NLP include Word2Vec, GloVe, BERT, and finally GPTx. With the development of word embeddings there was a significant shift towards using distributional semantics in language modelling. Models like Word2Vec (Mikolov, Chen et al., 2013; Mikolov, Sutskever, et al., 2013) and GloVe (Pennington et al., 2014) are unsupervised models that produce these embeddings by mapping words onto a high-dimensional semantic space. While Word2Vec accomplishes this based on neighboring context words, Glove relies on the global word co-occurrence matrix of words in a corpus. Word2Vec is a predictive model that generates word embeddings based on the distributional semantics of the words in a corpus. It utilizes a simple neural network to learn high dimensional word embeddings from a raw text, this can be achieved through training a binary classifier to predict a target word using its context words with Continuous bag-of-words (CBOW) model or the context given a word with Skip-gram models. With these models, weights were updated at each iteration to maximize the prediction, and producing an n-dimensional embedding matrix where each word in a vocabulary is represented as an embedding vector, thus allowing mathematical operations to be performed on the word vectors to identify relationships (Young et al., 2018).

Word embeddings became popular because of their capacity to represent words in a distributed latent space. To extract higher-level features from constituent words or n-grams deep neural network (DNN), models were utilized in conjunction with word embeddings. These abstract features would then be used in various NLP tasks such as sentiment analysis, text generation, machine translation and question answering. Given their success in computer vision tasks, convolutional neural networks (CNNs; LeCun et al., 1998; Krizhevsky et al., 2017; Razavian et al., 2014) and recurrent neural networks (RNNs; Elman, 1990) have been employed for NLP tasks as well. While CNNs are good at extracting position invariant features, they are very data heavy models because they include a large number of trainable parameters. Moreover, CNNs struggle to preserve sequential order and also have difficulties modeling long-distance contextual information (Kalchbrenner et al., 2014, Yin et al., 2017). RNNs, on the other hand, are designed with the goal of processing sequential data. The term "recurrent" refers to its ability to perform the same task on each sequence element, thus the output depends on previous computations and results. Essentially, RNNs possess the ability to retain memory of past computations and leverage that information in the current processing. However, simple RNNs can encounter the vanishing gradient problem, making it difficult to tune and learn parameters in the earlier layers. To address this limitation, alternative variants like long short-term memory (LSTM; Hochreiter & Schmidhuber, 1997; Jozefowicz et al., 2015) networks and gated-recurrent networks (GRU; Cho et al., 2014) were introduced.

LSTM is one of the most utilized RNN variants for NLP tasks. LSTMs are capable of learning relationship between words that are not in close proximity and they alleviate the exploding or vanishing gradient issue by selectively retaining or discarding information through the use of gates (input, forget, and output). This structure determines what information to add or remove from the output of the network from the previous time step. LSTM based models are widely adopted for sequence to sequence mapping tasks and implemented by using an encoder-decoder framework. These models have shown exceptional performance in different applications including machine translation, text summarization, question generation and modeling human conversations. von Davier's (2018) approach to automated item generation was based on a LSTM-RNN trained with publicly available items from personality tests. While LSTMs have been highly effective in addressing the challenges of long-term sequential dependencies, they too may suffer from the vanishing gradient problem during training because they have a very complex architecture, many non-linear transformations, and gradients are propagated through time to update the parameters. When gradients become too small, the model cannot efficiently learn long-term dependencies and can fail to make accurate predictions.

One of the notable advancements in NLP is the introduction of transformer models by Vaswani et al. (2017). Transformer models utilize self-attention mechanisms to capture dependencies across all positions in an input sequence, allowing them to model long-term dependencies more efficiently than LSTMs. Another significant advancement of transformer models is that the model can process sequential data in parallel, reducing resources needed for model training. Additionally, transformer models do not contain recurrence or convolution, therefore, cannot determine the order of a given sequence. Instead, the model incorporates a positional encoding mechanism to provide information about the absolute or relative positions of tokens in the sequence. These features, along with the availability of large training data, allow transformer models to achieve state-of-the-art performance on a wide range of NLP tasks. Several pre-trained models have been proposed that demonstrate the capabilities of transformer model implementations. These pre-trained large language models can be fine-tuned on task-specific datasets, and the model weights are adjusted to improve its representation of the specific corpus. Some notable examples of transformer models include BERT, and GPT; both were released in unidirectional and bidirectional variants. BERT is particularly notable for its ability to be pre-trained on large amounts of text data and then fine-tuned on specific downstream NLP tasks, leading to state-of-the-art results on many NLP tasks. Other important transformer-based models include Text-to-Text Transfer Transformer (T5; Raffel at al., 2020) that frames the tasks as a unified text-to-text transformation problem.

Building upon the success of GPT, OpenAI introduced GPT-2 and GPT-3, which are even larger models with significantly improved performance. GPT-2 was introduced in 2019 as a larger and more powerful version of GPT and was trained on over 8 million web pages and had 1.5 billion parameters (Radford et al., 2019). It was one of the largest language models at the time and demonstrated impressive performance, but it mostly required fine-tuning for every new task. von Davier (2019) used GPT-2 and fine-tuned the model with a large corpus of medical articles with the goal of generating medical licensing items. The challenges associated with collecting a large supervised training dataset for each new task, along with the potential of exploiting spurious correlations in training data with model expressiveness and narrowness of the training distribution, resulted in the need for larger models (Brown et al., 2020). GPT-3 was developed for this purpose with 175 billion parameters and trained using about 45 TB of text data from various sources. GPT-3 has generated widespread interest due to its ability to perform a wide range of NLP tasks with minimal fine-tuning and without any task-specific architecture modifications. Subsequently, OpenAI released a number of derivative models, including InstructGPT-3 (Ouyang, 2022), and Codex 12B (Chen et al., 2021). These models are commonly referred to as GPT-3.5 series and are comprised of the models trained on text and code from before the last quarter of 2021 (OpenAI, 2022). In this research, we will refer to all GPT-3 model variants including GPT-3.5 series as GPTx. Recently, OpenAI released a chatbot app, ChatGPT, utilizing GPT 3.5 series. ChatGPT uses reinforcement learning from human feedback to align the large language model with the user's intent (Ouyang, 2022). Within two months after its release it reached 100 million active users.

**Applications of Large Language Models in Education**

The development of GPT-3 and its variants has inspired new research into its potential applications in various areas. GPT-3 model and its successive iterations have also been utilized in educational research such as automated question generation (e.g., Raina & Gales, 2022; Settles et al., 2020; Wang et al., 2022), automated scoring (e.g., Mizumoto & Eguchi, 2023, Wu et al., 2023), generating educational content (e.g., Hocky & White, 2022; Moore et al., 2022; Walsh, 2022), and providing automated feedback for students (Zong & Krishnamachari, 2022).

Previous studies have also investigated other NLP-based models for generating reading comprehension questions and passages for educational assessment. For example, Shin (2021) combined template-based and non-template-based techniques to generate reading inference test items given reading passages based on the Harry Potter series. Fung et al. (2020) introduced a web-based system for generating reading comprehension questions and multiple-choice questions on grammar from a given text utilizing the T5 model. Similarly, Narayan et al. (2020) utilized another transformer-based model BERT to generate test items that could be answered using the information from the given reading passage directly.

While there has been research on generating reading assessment items, such as multiple-choice or cloze type questions with large language models like GPTx, there has been comparatively little attention on generating reading assessment passages. Recently, the Duolingo English Test (DET) employed GPT-3 to generate reading passages for both expository and narrative texts which represent the types of text that university students typically encounter in the target language domain (Attali et al., 2022). They also generated accompanying comprehension items and distractors for multiple-choice questions using GPT-3 for the generated passages. Duolingo English Text adopted "human-in-the-loop AI" throughout the process for test security, generation of test items, and scoring for test taker responses automatically (Burstein et al., 2021). Here, "human-in-the-loop AI" describes a process that includes an evaluation of the generated text by assessment researchers and designers. They evaluate new item types by assessing the degree to which they can facilitate item generation and what type of human intervention is necessary. In the case of item generation, a human review process was used to evaluate the fairness and possible bias in the automatically generated test passages and items (Burstein et al., 2021).

      Even though the applications of GPT-3 are not well explored within the reading assessment framework specifically, it has been utilized for various writing tasks in different domains. Gwern (2023) conducted extensive experiments with GPT-3 in fiction and non-fiction writing. For fiction writing, GPT-3 was able to produce coherent and creative stories, but longer outputs tended to become repetitive or lack originality. For non-fiction writing, GPT-3 showed promise in generating summaries of articles or research papers and generating new text based on given prompts. However, due to the potential for errors in technical or specialized content, human evaluation and editing were necessary. Moreover, Shakeri et al. (2021) utilized GPT-3 to develop a web application that allows people to write collaborative stories. Their results suggest that storytelling using GPT-3 can encourage roleplay and enhance creativity in the collaborative

writing process. Along the same lines, Noy and Zhang (2023) examined the productivity effects of using ChatGPT on mid-level professional writing tasks. Their results showed that ChatGPT substantially increased productivity and self-efficacy, and improved output quality and job satisfaction while reducing the inequality between workers. The various applications of GPT family models in text generation prompted this research on generating automated passages for a large-scale reading assessment of fourth graders.

## Methods

PIRLS provides comparative data on reading achievement at the primary school level across countries on a regular five-year cycle over 20 years. Reading literacy is fundamental to academic success and personal growth and PIRLS provides an invaluable instrument for assessing the impact of new or revised policies on reading achievement (Mullis, & Martin, 2019). The passages presented in the PIRLS assessment cover two purposes of reading that students encounter both in and out of school, those are *literary experience* and *acquire and use information*. In PIRLS, each reading purpose is linked to a particular type of text, such as literary experience is associated with reading fiction, and acquiring and using information is associated with informative articles and instructional texts. Following the structure in PIRLS, this research implemented the automated generation for both informational and literary passages.

Since its inception in 2001, after each PIRLS cycle, some of the reading passages have been released for public or academic use. For the last two cycles, the International Association for the Evaluation of Educational Achievement (IEA) no longer releases passages for public use without permission. Still, users can request access to the restricted use passages through their website (TIMSS & PIRLS, 2017). We constructed a pool with 24 passages that have been released over four cycles of PIRLS assessment from 2001 to 2016 to be used in prompts for automated reading passage generation with GPTx. In this paper, we specifically focused on analyzing narratives created utilizing three restricted-use passages, namely two informational passages: "Antarctica: Land of Ice" and "Ants," and one literary passage "Brave Charlotte." The original PIRLS passages are available upon request from IEA.

In this study, we utilized OpenAI's text-davinci-002 model, also referred to as InstructGPT, released in January 2022 as an updated and fine-tuned version of GPT-3 models, as mentioned above we refer this model as GPTx. To generate passages, we utilized Python (van Rossum & Drake, 1995) to send API requests with prompt design and parameters using text-davinci-002 text completion API. The associated cost for utilizing OpenAI's API service with text-davinci-002 was $0.02 for 1000 tokens.

By using natural language prompts and task demonstrations as context, GPT-3 is capable of effectively performing a diverse set of tasks with just a few examples, all without having to update the parameters of the underlying model (Gao et al., 2020). Consequently, prompt design is essential for large language models like GPTx to generate the desired response because it directly affects the quality and relevance of the output produced by the model. In this research, for each task, we generated passages using GPTx with first "zero-shot" learning, where only an instruction in natural language is provided without any demonstration, second "one-shot" learning, where a single demonstration together with a prompt containing an instruction is given to the model. In addition to the text prompts, we included the target audience's age as a part of the prompt for both one-shot and zero-shot scenarios. This additional information is provided to the model to ensure that the generated passages are age-appropriate, as they are intended for use in a fourth grade assessment. An example prompt for each scenario is given in Appendix A. We first generated outputs with one-shot learning (Figure 1A), after determining the topic of the story from the first set of outputs, we utilized that in the prompt of both one-shot (Figure 2A) and zero-shot (Figure 3A) learning for more detailed prompts.

Along with the prompts, we also manipulated the "temperature" setting in GPTx. This can be understood as a hyperparameter of the model that controls the randomness and variability of the generated output. For lower temperatures, GPTx chooses words with a high probability of occurrence, thus resulting in a more predictable and conservative output. When a higher temperature is chosen, the model explores a wider range of words resulting in more random, diverse and what some would call more creative output. Therefore, higher temperatures are recommended to generate ideas and stories. We generated stories with $t = \{0.5, 0.7, 0.9\}$ temperatures to allow for a range of outputs.

We generated informational and literary PIRLS stories, using the prompts and parameters mentioned above, with ten replications for each setting. After generating the passages, the selection mechanism involved first calculating the text difficulty score using an online text difficulty analyzer (Cathoven, 2023) associated with each generated story. This score is similar to a Lexile score (Stenner, 2022) that indicates the level of reading difficulty by combining measures of semantics and syntax that are represented by word frequency counts and sentence length, respectively. After calculating the text difficulty scores for both the original PIRLS passages and the generated passages, only the generated passages that fell within one standard deviation of the original passages' text difficulty scores were selected for further evaluation. Once selected passages were determined, a human editor went over the passages and corrected them for any grammatical and factual errors. The process involved three human editors, and they independently assessed the suitability of the content and provided suggestions for improvement by assessing the coherence, clarity, and logical consistency. A similar approach of human post-generation QC was taken in prior research (von Davier, 2018, 2019; Attali et al., 2022). The factual error checking was especially important in the case of informational passages because the model was prone to generating erroneous outputs in some of the replications, especially for certain statistics and numbers such as average rainfall in the Amazons or the average number of eggs laid by a queen bee. The finalized AI-generated passages are given in Appendix B.

To evaluate the appropriateness of generated passages, we compared original PIRLS passages with the GPT-3 generated ones within an online survey similar to the approach taken by von Davier (2018). We paired original passages with the AI-generated ones and devised three different online questionnaires with items using a 4-point Likert scale. After giving each passage following questions were asked: "The story is written at an adequate reading level for a fourth grader," "The story is written in a coherent manner," "Children will be able to identify the main topic of the story," "There are confusing or distracting elements in the story," and "This story can engage children to answer questions." We recruited 150 human expert raters (50 for each survey) through Amazon Mechanical Turk, the participants were awarded $1.50 upon satisfactory completion of the survey. We set up additional qualifications for the expert rater selection. To be selected as rater of passages of the study, they must have a Bachelor's degree and be working in the education industry. Additionally, we used different measures to identify and exclude potential inattentive users and non-respondents (bots) in Amazon MTurk. To identify bots, we increased the time between Human intelligence task (HIT) completion and auto-approval to examine the data before approving or rejecting the HITs. Moreover, we rejected HITs with unreasonable response times, this was determined using the median absolute deviation (MAD) statistic on the overall completion time. This type of outlier detection (Leys et al., 2013) was shown to be successful in the detection of differential item functioning (von Davier &

Bezirhan, 2022) and can be readily applied to the detection of suspiciously short response times. The last measure was to incorporate an attention question in each survey to sort out careless respondents. This resulted in a final sample of 50 respondents for each survey.

## Results

For the informational PIRLS passages "Antarctica: Land of Ice" and "Ants", with one-shot and zero-shot learning along with the additional age/grade indicator, we generated "The Amazon: Green Lungs of the Planet" and "Bees." Overall, for one passage 160 different passages were generated, 40 for each condition, and their text difficulty scores were calculated. In the generation process, we first used one-shot prompting, shown in Figure 1A in Appendix A, and let GPT-3 generate passages that are similar to the given example passage. This first batch with 10 replications was to determine the topic of the generated passage and therefore discarded. GPTx almost always generated the passages around the same topic, for example, for "Ant"s almost all of the generated passages were about bees.

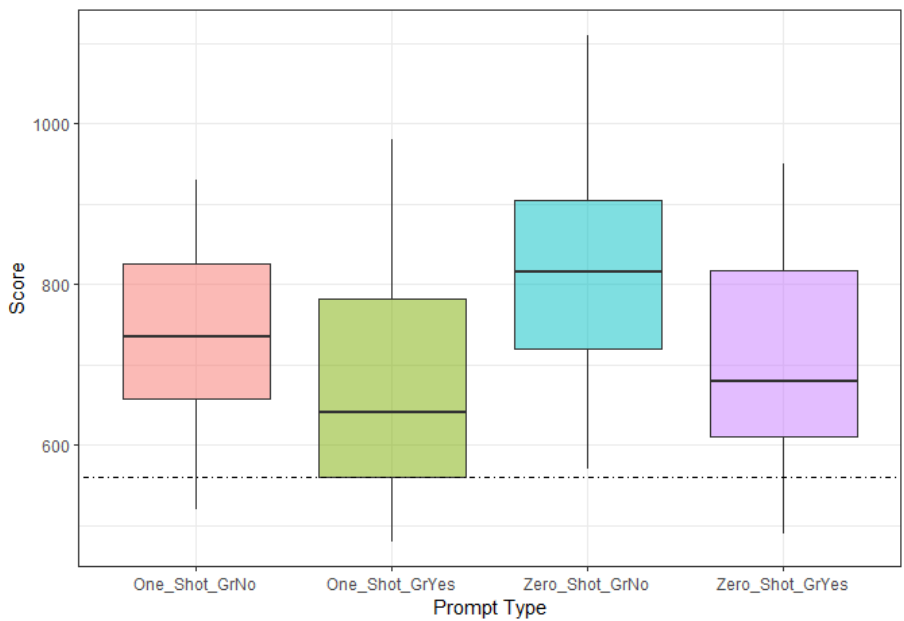

Figure 1. Text difficulty scores for generated "Bees" passage under different prompting conditions.

Figure 1 shows the distribution of text difficulty scores for the "Bees" passage for one-shot or zero-shot learning, with either grade information included or not. The "Bees" passage was based on the "Ants" passage in PIRLS which had a text difficulty score of 560, the horizontal line in Figure 1 represents this value. From all the prompt types, passages from one shot learning with grade information included had a lower text difficulty score compared to the other generated passages. There is no clear pattern regarding the variance associated with the prompt types.

The other informational passage "The Amazon: Green Lungs of the Planet" was prompted using the PIRLS "Antarctica: Land of Ice" passage as an example. It has to be pointed out that these generated passages are not simple copies where one word is replaced by another word to vary content. Unlike traditional AIG approaches (Gierl et al., 2020), the GPTx-generated passages are based on priming a large language model with a context, and then have the model generate an independent text, inspired by requesting a type of text, a topic, or by providing an example. The calculated text difficulty scores are presented in Figure 2, the horizontal line shows the score for the "Antarctica: Land of Ice" passage. For both one-shot and zero-shot prompting conditions, inclusion of grade/age information reduced the text difficulty score for the generated passages.

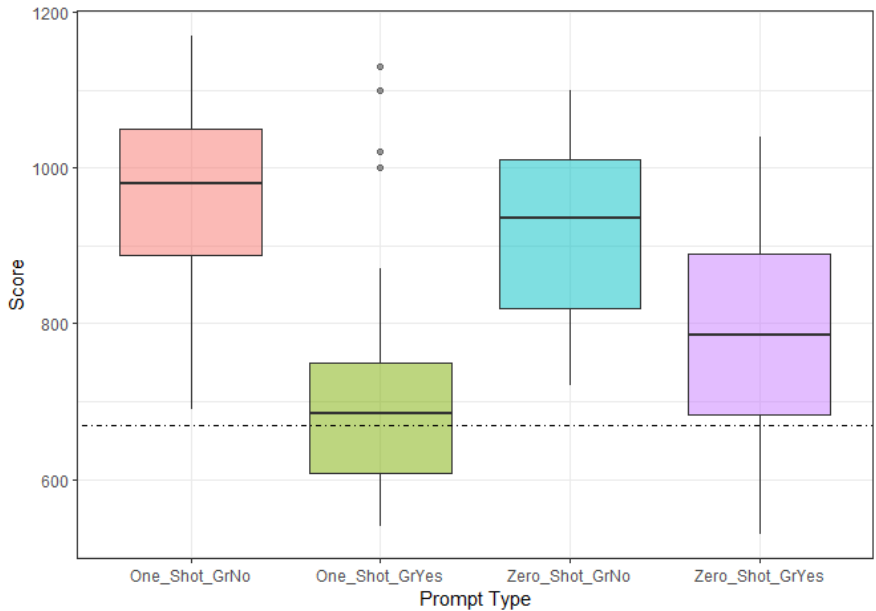

Figure 2. Text difficulty scores for generated "The Amazon: Green Lungs of the Planet" passage under different prompting conditions.

For the literary passage generation, the prompting approach was slightly different than the one used for generating informational passages. Similar to the informational passage generation, the first step was to generate ideas about a story utilizing an existing PIRLS passage. "Brave Charlotte" was used as a prompt for the literary passage generation, and GPTx came up with various different storylines. "Brave Charlotte" is about a brave sheep which helps the shepherd in a difficult situation. The generated stories were about the importance of friendship, kindness, and community, which were not very different from the general theme of the prompted story. Utilizing one of these ideas, we started generating "Coco the Rabbit." While generating literary passages with GPTx, we faced the challenge of premature stopping points, which made it difficult to generate longer stories. To overcome this issue, we adopted a stepwise approach and leveraged previously generated text as prompts, in addition to providing instructions to generate full-length stories successfully.

The finalized stories were used in our expert judgment exercise collection along with the PIRLS passages. To prevent the experts from being affected by similarities between the prompted and generated passages, we paired the informational passages "Bees" with "Antarctica: Land of Ice" and the "Amazons: Green Lungs of the Planet" with "Ants." However, we only paired literary passages together as the two stories had different main storylines and characters, and thus were not directly comparable.

In this research, we employed a four-category Likert scale to measure participant attitudes toward each passage. We gathered answers from a sample of 50 expert judges for each passage. Figure 5 displays the distribution of responses and level of agreement and disagreement among judges for GPTx generated "Bees" passage and PIRLS original "Ants" passage. The results indicate that a high portion of judges agreed or strongly agreed that both passages demonstrated similar levels in the assessed qualifications. In particular, 92% of the judges agreed or strongly agreed with the adequateness of the GPTx generated passage. In comparison, 96% of the judges agreed strongly that original PIRLS passage is adequate for the target population of students. Other questions exhibited a similar trend, with marginal variations between 2% and 8%. Positive evaluations towards engagement were observed for both passages, with the PIRLS passage scoring slightly higher (86%) than the GPTx generated passage (84%). The coherence of the passages revealed the largest gap in agreement with 94% of the participants found the PIRLS passage coherent compared to 84% for the AI-generated passage.

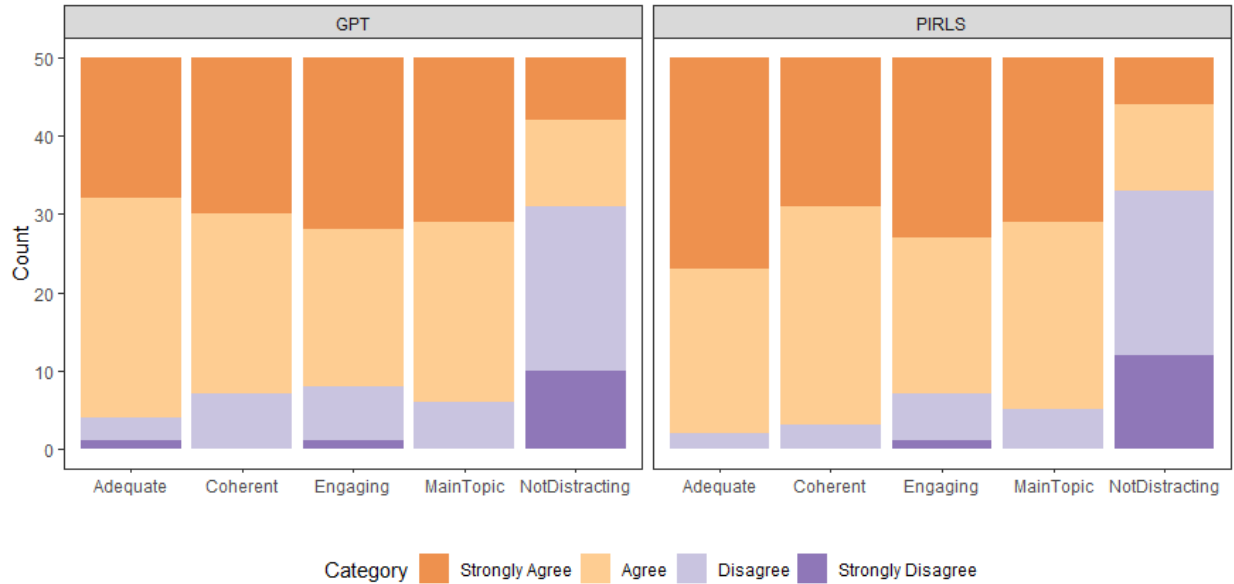

Figure 3. GPT generated "Bees" and PIRLS original "Ants" passage survey results

Similar outcomes were observed for comparing the "Antarctica: Land of Ice" and "Amazons: Green Lungs of the Planet" passages. The distribution of responses and level of agreement and disagreement among participants for these passages are given in Figure 4. Overall, the agreement and disagreement levels aligned well between the AI-generated and the original human-generated PIRLS passages across all questions. However, we observed a larger difference between the individual response categories for certain questions.

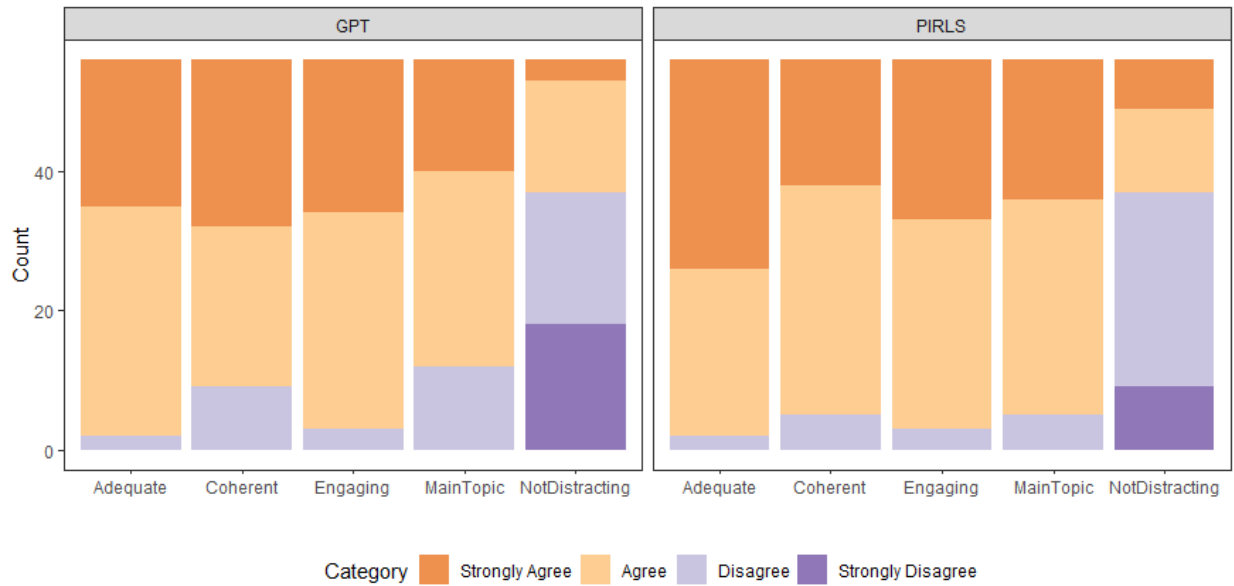

Figure 4. GPT generated "Amazons" and PIRLS original "Antarctica" passage survey results

For both "Amazons" and "Antarctica" passages, 97% of participants agreed that the passages were adequate for fourth grade readers. However, the strongly agree category was higher for the original PIRLS passage (54%) compared to the AI-generated one (38%). For coherence, the positive attitude towards the original PIRLS passage (91%) was higher compared to GPTx generated passage (84%), this suggests that human-generated passages may have been more logically structured and easier to follow than the GPTx generated one. However, more judges strongly agreed (43%) with the statement for the GPTx generated passage compared to the original PIRLS passage (32%), when we look at the individual categories for coherence. Moreover, identifying the passage's main topic was deemed harder for GPTx generated passage, with 79% of the participants agreeing with the statement, compared to the agreement with the human-written passage (91%). Finally, the overall agreement on the passage not being distracting was similar for both passages, but more judges had a stronger evaluation towards GPTx generated passage (32%) being more distracting than the original PIRLS passage (16%).

However, when taking strongly agree and agree categories together, there was no difference observed between the general level of agreement for AI generated (66%) vs. human generated passage (66%).

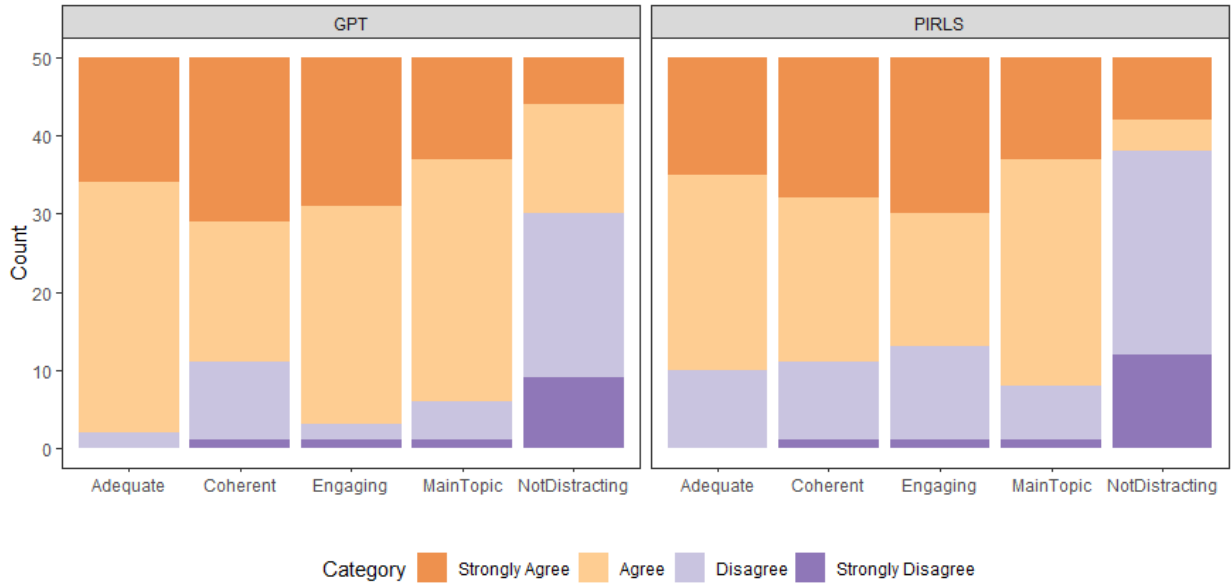

Figure 5. GPT generated "Coco the Rabbit" and PIRLS original "Brave Charlotte" passage survey results

Lastly, Figure 5 displays the level of agreement and disagreement between "Coco the Rabbit" and "Brave Charlotte" stories. A somewhat similar pattern in the agreement was also observed with these literary passages, with GPTx generated passage being slightly more adequate and engaging and less distracting compared to the original PIRLS passage. For the passage being adequate for the fourth graders, judges agreement was higher for AI-generated passage (96%) compared to the original PIRLS passage (80%). A similar pattern was observed for the engagement, 94% of the participants agreed with the AI-generated passage to be engaging, whereas only 74% of the participants agreed about the same statement for the original passage. Lastly, about 16% more people found the human written passage more distracting compared to the GPT-generated passage.

**Discussion and Conclusion**

With the recent technological advancements, the field of NLP has been completely transformed by the emergence of large language models, and OpenAI's GPTx family of models has appeared as one of the most sophisticated and powerful large language models available today. Researchers and practitioners in numerous disciplines have taken notice of it due to its impressive capacity and performance. In this research, we investigated the text generation ability of GPT-3 in the context of an international large-scale reading assessment. We utilized GPT-3 to automatically generate literary and informational passages similar to those typically employed in the PIRLS assessment.

We experimented with two different prompt designs along with the age information of the audience to generate informational passages. After determining the topic of the specific passage more detailed prompts given in Appendix A were utilized. The results revealed that the one-shot learning, along with the specific information about grade, usually showed the best performance in terms of matching the text difficulty score of the original passage. This is in line with previous research that GPTx tends to perform better when more examples are provided along with clear instructions and this is interpreted as GPTx learning from the examples at runtime (Brown et al., 2020). Given that the generated passage was intended to be tailored towards a fourth grade assessment and compared according to their test difficulty scores, providing clear instructions about the intended audience was beneficial for both one-shot and zero-shot prompts. This demonstrates that the carefully constructed prompts primarily direct the model to access existing knowledge and perform the task, which is referred to as direct task specification (Reynolds, & McDonell, 2021).

Moreover, the results suggest that GPTx can produce passages that closely resemble those in the PIRLS assessment, not only in terms of length, vocabulary, and difficulty but also in adequacy, coherence, engagement, and ability to cause distraction. The empirical analysis results also revealed that in addition to making it more difficult to determine the main topic, AI-generated passages could also be more distracting for the readers. This is likely due to the fact that passages in the actual assessment were written by human authors for the purpose of the assessment. At the same time, those generated by GPTx may lack intentional organization and coherence, therefore, making it more difficult for the readers to identify the main topic. In contrast, for the literary story, the participants found the AI-generated passage to be less distracting and more engaging for the fourth grade readers than the original PIRLS passage. This finding may suggest that the iterative prompting approach might contribute to generating more appealing and easier-to-comprehend stories.

While GPTx generated passages demonstrated some promising results, a number of potential limitations need to be considered for future research. First of all, this research focused on demonstrating the feasibility and applicability of a GPT family model to reading passage generation. Given that recently more advanced models were released by OpenAI, future research should consider using ChatGPT or GPT-4 (OpenAI, 2023). Even though we experimented with some prompt designs in this research, it is quite limited to observe the full effect of different prompting techniques on age-adequate reading passage generation. Therefore, it would be worthwhile to explore more comprehensive prompt engineering approaches, including both manual techniques (e.g., Reynolds, & McDonell, 2021) and automatic methods (e.g., Gao et al., 2020) within the context of the reading assessment passage generation. Lastly, the sample size in the empirical analysis of expert judgements was relatively small which may limit the generalizability of the results.

    Nevertheless, this research demonstrates that GPT family models could be effectively utilized for automated passage generation in the context of a large-scale reading assessment. Considering the high costs and significant time investment associated with human-authored assessment development and the copyright concerns that often arise, large language models present a promising opportunity to streamline and enhance current practices in assessment development.

# Appendix

## A. Example Prompts

```
This is an informative story generator.
The story should have multiple parts and the sections should be informative
and engaging [for a 10-year-old]. An example story is given below.

Story: Ants
Small and Strong
Lift up a rock, and a family of ants might be crawling there.
Ants are small insects, but they are very strong. Ants have six strong legs
that help them carry big loads such as sticks and other insects. They can
lift 20 times their own body weight.

Building a Home
Most ants live in nests in the ground. Each nest is like an underground city.
It has rooms, called chambers, where the ants live and work.
The chambers are connected by tunnels.
..........
```

Figure 1A. Initial Prompt for Bees Passage

```
This is an informative story generator.
Generate an informative story about Bees [for a 10-year-old]. It includes sections about
bees' body, their honey production, social life and importance to ecosystem.
The sections should be informative and engaging [for a 10-year-old].

Story: Ants
Small and Strong
Lift up a rock, and a family of ants might be crawling there.
Ants are small insects, but they are very strong. Ants have six strong legs
that help them carry big loads such as sticks and other insects. They can
lift 20 times their own body weight.

Building a Home
Most ants live in nests in the ground. Each nest is like an underground city.
It has rooms, called chambers, where the ants live and work.
The chambers are connected by tunnels.
..........
```

Figure 2A. One-shot prompt for Bees Passage

```
This is an informative story generator.
Generate an informative story about Bees [for a 10-year-old]. It includes sections about
bees' body, their honey production, social life and importance to ecosystem.
The sections should be informative and engaging [for a 10-year-old].
```

Figure 3A. Zero-shot prompt for Bees Passage

## B. GPT Generated Passages

**Passage 1. Bees**

Bees are small, flying insects that are known for their role in pollination and for producing honey. They are found on every continent in the world except Antarctica.

Bees' Bodies

Bees have adaptations that help them collect nectar and pollen from flowers. For example, they have a long tongue that unrolls to drink in nectar. They also have specialized hairs on their body that collect pollen.

There are three sections to a bee's body: the head, the thorax, and the abdomen. The head of a bee contains the eyes, antennae, and mouthparts. The thorax is the middle section where the wings and legs are. The abdomen is the bee's back end and where the bee has the stinger. The female bees have a stinger that they use to defend themselves and their hive.

Honey

Bees produce honey by collecting nectar from flowers. Nectar is a sugary liquid that is produced by flowers. Bees use their long tongue to collect the nectar and store it in their special honey stomach. Once the bee returns to the hive, the nectar is mixed with enzymes and turned into honey. The honey is then stored in the honeycomb. When the weather is cold and no flowers are blooming, the bees eat the honey to stay alive.

Social Life

Bees are social insects that live in hives (or colonies). A bee colony is made up of three types of bees: the queen bee, the worker bee, and the drone bee. Each bee has a specific job to do in the colony. The queen is the biggest bee in the hive and the only bee that can lay eggs. She can lay 2,000 eggs a day in the spring or summer months. The workers are female bees, and they do all the work in the hive! They collect nectar, make honey, build the honeycomb, care for the eggs and young bees, and protect the hive. The drones are male bees, and their job is to mate with the queen. In the winter, the hive goes into survival mode, so the drones are kicked out.

Importance to Ecosystem

Bees are essential to the ecosystem because they help to pollinate plants. Pollination is the process of transferring pollen from the male part of the plant to the female part of the plant. This process is necessary for the plant to produce seeds. Bees are responsible for pollinating about one-third of the food we eat, including fruits, vegetables, and nuts. Bees are in danger because of the use of pesticides and the loss of habitat. You can help bees by planting flowers in your garden and avoiding the use of pesticides.

**Passage 2. The Amazons: Green Lungs of the Planet**

The Amazon: Green Lungs of the Planet

The Amazon is a large rainforest that takes up most of the Amazon Basin in South America. The basin is home to the Amazon River, the second longest river on Earth. The Amazon rainforest covers multiple countries and an area of nearly 6 million square kilometers. This is almost half of the total rainforest area on Earth! The Amazon rainforest is an important part of the world's climate because it helps to adjust the Earth's temperature. It also plays a vital role in the global water cycle, producing much of Earth's oxygen.

The Weather in the Amazon

The climate in the Amazon rainforest is very warm and humid. It rains a lot in the Amazon. The average rainfall each year is 300 centimeters. This makes 200 rainy days in a year! There are only two seasons in the Amazon: the wet season and the dry season. The wet season is from October to May, and the dry season is from June to September. It is the wet season when the Amazon receives most of its rain.

The Plants and Animals in the Amazon

There are different kinds of plants and animals in the Amazon rainforest. Many plants and animals in the rainforest are not found anywhere else in the world. Some of the animals that live in the rainforest are monkeys, jaguars, sloths, snakes, and toucans. The Amazon rainforest is also home to many different types of plants, such as the rubber tree, the Brazil nut tree, and the cocoa tree. Some trees are as tall as 40 meters, and you can use their leaves as an umbrella during a rainstorm!

You have a letter from a team of researchers studying in the rainforest and can learn more about the Amazons by reading their letter.

Dear Friend,

The Amazon is the world's largest rainforest. It is home to many different kinds of animals and plants. Unfortunately, Amazon is under threat. We are one of the research teams here in the

Amazon that works on threats to the rainforest. We study deforestation, climate change, and fires in the Amazon. We try to find ways to help protect the Amazon and its animals.

One of the biggest threats to the Amazon is deforestation. Deforestation is when trees are cut down, and the land is cleared for other uses, such as farming. Deforestation is a significant problem because it destroys the home of many animals and contributes to climate change.

We are working hard to understand the long-term effects of deforestation on the climate. We are also working to develop ways to regenerate the forest. We invite you to learn more about our research and to help us spread the word about the importance of the rainforest.

Thank you for your support,
The Amazon Research Team

**Passage 3. Coco the Rabbit**

The shiny button eyes of the little rabbit stared up at the sky. The rabbit's name was Coco, and she was very excited. She was going on her first adventure. She had never been outside her meadow before and was very curious about the world. Her friend, a little fox named Max, was going with her. Max had been to the forest many times, and he promised to show Coco around. Coco and Max hopped through the meadow, stopping to smell the flowers and chase the butterflies. They walked through the forest, stepping over roots and ducking under branches. They climbed over a fallen tree, and Max helped Coco down the other side. They walked and walked until they came to a clearing. In the clearing was a big, old house. It looked like it had been abandoned for years.
"This is where the adventure starts," Max said.

Coco's heart started to beat faster. She was scared, but she was also excited. She wanted to see what was inside the house.

Max and Coco crept up to the house. Max pushed open the door, and they both stepped inside. It was very dark and musty inside. They could see the dust floating in the air. Max and Coco coughed and covered their mouths with their paws.

"Let's go explore," Max said.

Coco followed Max as he showed her around the house. They went into different rooms and saw old furniture covered in dust. In one room, there was a big mirror. Max and Coco looked at themselves in the mirror. They both jumped when they saw a face appear in the mirror behind them. It was a scary face with big teeth.

Max and Coco screamed and ran out of the room.

They ran and ran until they were outside the house. They were both panting, and Coco's heart was beating so fast she thought it would come out of her chest.

"That was so scary!" she said. "But who was that?"

"I don't know," Max said. "But we're never going back there to find out!"

"Maybe we should go back," said Coco.

"No way," said Max. "That place is haunted."

Coco thought for a moment. She was a brave rabbit and wouldn't be scared of a little ghost.

"I'm going back," she said. "If you're too scared, then you can wait here."

Max didn't want to be left alone, so he followed Coco back to the house. They went into the room with the mirror, and sure enough, the ghost was there again. But this time, Coco wasn't scared.

"Who are you?" she asked.

"I'm the ghost of the house," the ghost said. "I'm here to protect this house."

"But why are you so scary?" Coco asked.

"I'm not trying to be scary," the ghost said. "I just want to make sure no one steals anything from this house."

"We're not going to steal anything," Coco said. "We're just exploring."

"I'm sorry," the ghost said. "I didn't know that. You can explore all you want. Just be careful not to break anything."

But Coco had an idea, and if he was the ghost of the house, maybe he could show them around.

And so, he did. He showed them all the hidden passages and secret rooms. He even showed them where the treasure was hidden.

"This has been the best adventure ever!" Coco said when they were finished. "Thank you for showing us around."

"You're welcome," the ghost said. "And thank you for being kind. I'm not used to people being kind to me."

"That's what friends are for," Max said. "That's what we do for each other."

They said goodbye to the ghost and left the house. Coco thought to herself; if you give a chance to get to know someone, even a ghost, you might find they're not so different from you after all.